\PassOptionsToPackage{table}{xcolor} 
\documentclass[10pt, a4paper]{article}
\usepackage{booktabs}
\usepackage[tone]{tipa}
\usepackage[final]{lrec2026} 

\usepackage{listings}
\usepackage{enumitem}
\usepackage{xurl}
\usepackage{textcomp}

\definecolor{codegreen}{rgb}{0,0.6,0}
\definecolor{codegray}{rgb}{0.5,0.5,0.5}
\definecolor{codepurple}{rgb}{0.58,0,0.82}
\definecolor{backcolour}{rgb}{0.95,0.95,0.92}

\lstdefinestyle{mystyle}{
    backgroundcolor=\color{backcolour},   
    commentstyle=\color{codegreen},
    keywordstyle=\color{magenta},
    numberstyle=\tiny\color{codegray},
    stringstyle=\color{codepurple},
    basicstyle=\ttfamily\footnotesize,
    breakatwhitespace=false,         
    breaklines=true,                 
    captionpos=b,                    
    keepspaces=true,                 
    numbers=left,                    
    numbersep=5pt,                  
    showspaces=false,                
    showstringspaces=false,
    showtabs=false,                  
    tabsize=2
}

\title{Diacritic Restoration for Low-Resource Indigenous Languages: Case Study with Bribri and Cook Islands M\={a}ori}

\name{Rolando Coto-Solano\textsuperscript{1}, Daisy Li\textsuperscript{1}, Manoela Teleginski Ferraz\textsuperscript{1}, Olivia Sasse\textsuperscript{1}, \\ {\bf \large Cha Krupka\textsuperscript{1}, Sharid Lo\'{a}iciga\textsuperscript{2} and Sally Akevai Tenamu Nicholas\textsuperscript{3}} }

\address{\textsuperscript{1} Dartmouth College,\\
\textsuperscript{2} University of Gothenburg, Department of Philosophy, Linguistics and Theory of Science\\
\textsuperscript{3} The University of Auckland (Waipapa Taumata Rau) \\
         \{rolando.a.coto.solano, daisy.li.26, manoela.e.teleginski.ferraz.27, \\ olivia.g.sasse.25, cha.j.krupka.25\}@dartmouth.edu, sharid.loaiciga@gu.se, \\ ake.nicholas@auckland.ac.nz\\}

\abstract{
We present experiments on diacritic restoration, a form of text normalization essential for natural language processing (NLP) tasks. Our study focuses on two extremely under-resourced languages: Bribri, a Chibchan language spoken in Costa Rica, and Cook Islands M\={a}ori, a Polynesian language spoken in the Cook Islands.  Specifically, this paper: (i) compares algorithms for diacritics restoration in under-resourced languages, including tonal diacritics, (ii) examines the amount of data required to achieve target performance levels, (iii) contrasts results across varying resource conditions, and (iv) explores the related task of diacritic correction. We find that fine-tuned, character-level LLMs perform best, likely due to their ability to decompose complex characters into their UTF-8 byte representations. In contrast, massively multilingual models perform less effectively given our data constraints. Across all models, reliable performance begins to emerge with data budgets of around 10,000 words. Zero-shot approaches perform poorly in all cases. This study responds both to requests from the language communities and to broader NLP research questions concerning model performance and generalization in under-resourced contexts.
\\ \newline  }

\begin{document}

\maketitleabstract

\section{Introduction}


Building digital corpora is essential to preserve invaluable linguistic materials. Language documentation provides the source data for many subfields of linguistics, such as phonology, morphology, syntax, semantics, and pragmatics \citep[p.~2]{Chelliah2021}. Over the past decade, many researchers, linguists, and consortiums have worked closely with native speakers and language communities to create
corpora including digitized text, audio, transcriptions, translations, stories, etc.  \citep[p.~89]{agarwal-anastasopoulos-2024-concise}. In the best case scenario, these corpora can be used not only to preserve, but also revitalize endangered languages \citep{shi-etal-2021-highland} and reclaim Indigenous sovereignty \citep{viatori-ushigua-speaking-sovereignty, leonard2023refusing}. High-quality corpora can also be used to train NLP language education tools for learners \citep{zhang2022nlphelprevitalizeendangered}.

In many Indigenous communities, writing is of very recent adoption, and there is considerable variation on how words are spelled. This is a natural process which has occurred in every written language \citep{pombo2012, moessner2017standardization}, and it is not usually an issue for communication. However, it is an issue for the automatic processing of the data, since variation can hinder performance \citep{pettersson2013normalisation, salloum2011dialectal}.


This is especially complicated with diacritics and non-alphabetic characters. Diacritics are symbols that are appended to consonants or vowels. They can indicate tone (e.g. the macron in the Mandarin \textit{s\={a}n} `three'), stress (e.g. the accute accent in Spanish \textit{caminé} `I walked'), nasality (e.g. the Polish ogonek in \textit{kąt} `angle'), and different vocalic qualities (e.g. the German umlaut in \textit{\"{o}l} `oil') amongst their many uses. These can show even greater variation in Indigenous language data \citep{roberts2013tone}. They also elicit a wide range of opinions of the users of these languages: While school teachers and linguists insist on using diacritics because they feel they bring precision to the text, poets and other writers sometimes think the diacritics make the text feel crowded or difficult to read \citep{bird1999marking, roberts2009visual,kadyamusuma2013potential}, or simply do not use them because they feel the text can be understood without them \citep{kishindo1998standardization, bernard2002does,saynes2002zapotec,koffi2014towards}. This variation in opinions leads to variation in spelling, which compounds to the difficulties of building corpora.

Diacritics are not just important for a standard representation of text, they also make a difference in their processing using NLP tools. The same diacritic can be encoded in multiple ways in Unicode, resulting in variation. Most state-of-the-art large language models (LLMs) use preprocessing steps to strip diacritics that might be inconsistently encoded or inconsistently omitted in certain scripts. However, stripping diacritics (or failure to normalize diacritics) leads to degradation in the performance of downstream tasks such as POS-tagging or dependency parsing. This degradation is exacerbated in scripts such as Hebrew, which rely heavily on diacritics to disambiguate. Thus, the preferred method of handling diacritic inconsistencies is to apply Unicode normalization using any standard library module \citep{gorman-pinter-2025-dont}.

In this paper, we present experiments on diacritic restoration for two extremely under-resourced languages: Bribri and Cook Islands M\={a}ori. We evaluate several approaches, including character-based translation models, fine-tuned LLMs and zero-shot LLMs for the task. Our results show that character-based models work best, and that data masses as low as 10000 words can produce acceptable restoration models. This paper not only benchmarks the well-established task of diacritic restoration for two extremely under-resourced languages, but also responds to the communities' request for the development of NLP tools. In addition, we provide a detailed linguistic analysis of the outputs, with particular attention to tones and glottal stops, known to be problematic for Deep Learning and critical to work of language documentarians.

Our contributions are as follows: We:

\begin{itemize}[noitemsep]
\item compare the performance of different algorithms when restoring diacritics in low-resource languages;
\item analyze how much data is needed to reach specific performance levels; 
\item contrast the diacritic restoration outcomes between high and low resource languages; and 
\item conduct a brief study of the related task of diacritic correction.
\end{itemize}

\section{Related Work}\label{sec:relatedwork}

\paragraph{Diacritic Restoration}
Some of the research on diacritic restoration originates from work on normalization of historical or non-standard varieties of a language \citep{munoz-ortiz-etal-2025-evaluating, ljubesic-etal-2016-corpus}. Other studies focus on heavily-diacritized languages such as Arabic, where diacritics are often omitted in written text \cite{masmoudi2019automatic,alqahtani2020multitask,hifny2021recent}. Broadly, two main approaches have been proposed: statistical or rule-based models, and deep learning models. Statistical models are typically more computationally efficient \citep{hifny2021recent}, but recent research has found that deep learning models are increasing their accuracy in under-resourced language tasks \citep{naplava-etal-2018-diacritics}. Diacritic restoration has also been investigated in European languages such as Hungarian, Turkish, and South Slavic languages \cite{novak2015automatic,ozer2018diacritic,hucko2018diacritics,stankevivcius2022correcting}, and in tonal languages like Vietnamese \citep{nga2019deep, tran2021study, le2021diacritics}, Igbo \citep{ezeani2016automatic} and other African languages such as Yoruba and Kikuyu \citep{depauw2007automatic}.

Interestingly, there is work on diacritic restoration in te reo M\={a}ori, a Polynesian language spoken in Aotearoa New Zealand. \citet{cocks2011word} used a corpus of 3.8 million words and a Naïve Bayes classifier to obtain a 99\% acccuracy when reconstructing diacritics, and \citet{nakano2023using} used an RNN-based algorithm and a 463,000 word corpus to obtain a 96\% accuracy.

\begin{figure}[!ht]
\begin{center}
\includegraphics[width=\columnwidth]{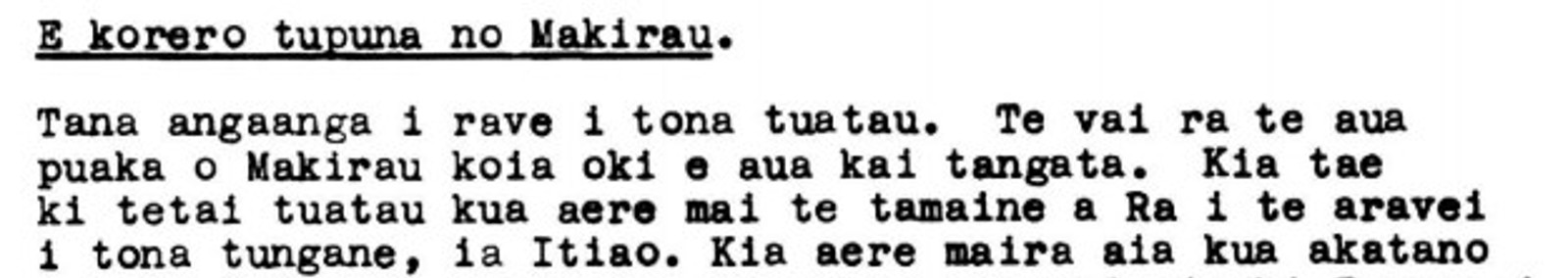}
\caption{Example of typewritten text in CIM with long vowel macrons and glottal stop saltillos absent \citep{Ponga:1975}. In the \citet{Nicholas2017manual} orthography, the first words would be \textit{T\={a}na {\textquotesingle}anga{\textquotesingle}anga} `her work'.}
\label{fig1-typewritten-cim}
\end{center}
\end{figure}

Cook Islands M\={a}ori is a Polynesian language spoken in the Cook Islands by 12,500 people, plus 10,000 in the diaspora \citep{statscook, nicholas2018}. It is an endangered language, which means that there is a decreasing number of children who speak it. It is closely related to other Polynesian languages like te reo M\={a}ori, Hawaiian and Tahitian. In this language, henceforth referred to as CIM, there is a partially accepted system for spelling which includes two diacritics \citep{Nicholas2017manual}, but most public text is written with few or no diacritics. Figure \ref{fig1-typewritten-cim} shows an example of such text\footnote{We want to emphasize that orthographic variation is not an `error' or a `mistake' of the part of the speaker of the Indigenous language. Our goal is not to eliminate variation, but to respond to two practical considerations: (i) the community’s request for the development of NLP tools, and (ii) the existence of certain orthographic standards that the community itself has adopted. Orthographic variation should never be treated as a mistake, and we encourage writers to continue producing text freely, in their preferred forms. The responsibility for handling standardization lies with the developers of NLP tools, not with language users. At the same time, computer scientists and NLP practitioners must take care not to establish a single variant as the de facto norm, as doing so risks erasing valuable linguistic diversity, an issue that deserves explicit discussion within the field.}. In order to make it compatible with diacritically-marked text, the diacritics would need to be added, a time-consuming task that requires expert knowledge which few people possess.


\paragraph{Orthographic Correction and Diacritics}

Another way in which variation is expressed is in divergent and idiosyncratic uses of orthography. This is common during the normalization of orthographic norms, but it adds difficulty to NLP processing.

\begin{figure}[!ht]
\begin{center}
\includegraphics[width=\columnwidth]{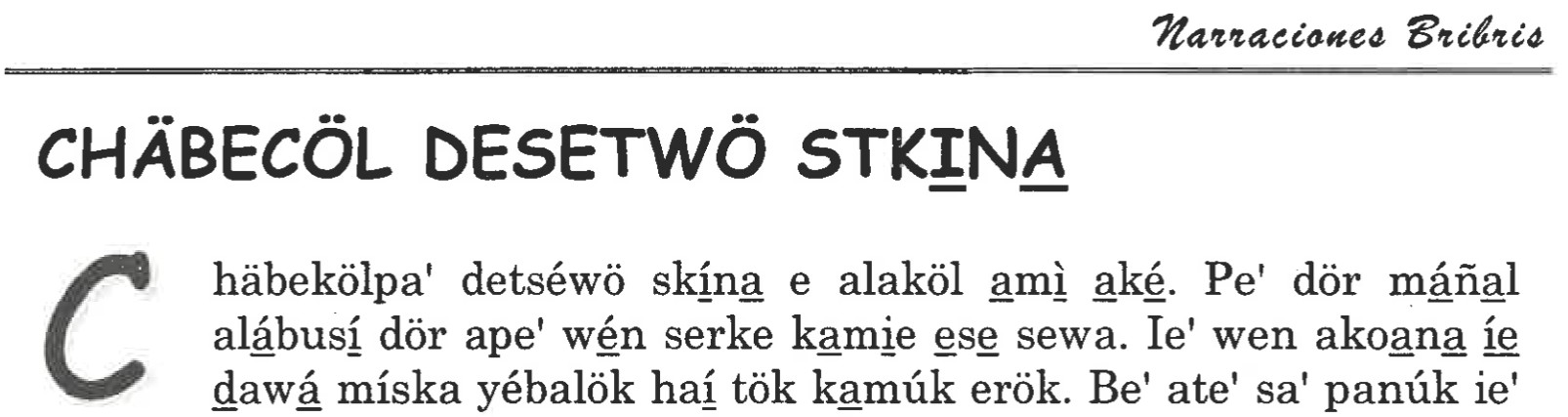}
\caption{Example of spelling variation in Bribri \citep[41]{ietsay-narraciones}. Using Constenla's \citeyearpar{cursobribri} orthography, the first two lines would be: \textit{Tkab\`{\"{e}}k\"{o}l dits\`{e}w\"{o} tsk\`{i}n\underline{a}: Tkab\`{\"{e}}k\"{o}l dits\`{e}w\"{o} tsk\`{i}n\underline{a} e' al\'{a}k\"{o}l \underline{a}m\underline{\`{i}} ak\`{\"{e}}}. ``The Snake People are Created: The Snake People who were created (came from) a woman, (her) mother (and her) brother". For example, the word \textit{dits\`{e}w\"{o}} `clan, people' appears as \textit{desetw\"{o}} and \textit{dets\'{e}w\"{o}}.}
\label{fig2-printed-bribri}
\end{center}
\end{figure}

Bribri is a Chibchan language, spoken by 7,000 people in southern Costa Rica \citep{censo2011}. It is also vulnerable \citep{carlosvitalidad}, and it has at least three different orthographic standards, represented in \citet{cursobribri}, \citet{gramaticajara} and \citet{margery}. Text produced by other authors and by online users shows considerable idiosyncratic variation, as shown in figure \ref{fig2-printed-bribri}. This doesn't need to be corrected to be understood by humans, but it might need to be standardized to be used along other NLP tools.

\paragraph{Cook Islands M\={a}ori: Long vowels and glottal stops}

The Cook Islands M\={a}ori Revised New Testament   orthography (henceforth CIMR) \citep{Nicholas2017manual, CIMR2014} 
for CIM uses one diacritic and one non-alphabetic symbol, shown on Table \ref{tab:diacritics-cim}: A macron to mark long vowels, and the glyph saltillo (A78C) to indicate a glottal stop phoneme. This last one shows considerable variation, and regularly appears as an apostrophe (0027), a left quote (201C) or an `okina (02BB).

\begin{table}[h!]
\centering
\begin{tabular}{ll}
\toprule
Long vowel (macron)     & \={e} \\
Glottal stop (saltillo) & \textquotesingle e \\ 
\bottomrule
\end{tabular}
\caption{Diacritics and special symbols in CIM}
\label{tab:diacritics-cim}
\end{table}

\paragraph{Bribri: Tones, nasals and glottal stops}

Bribri uses three diacritics and one non-alphabetic symbol to indicate tone, an underline diacritic for vowel nasality, and an umlaut for lax vowel quality. These are shown in Table \ref{tab:diacritics-bribri}. In this paper we will ignore consonantal diacritics and instead consider them part of the consonant, in particular the tilde in `ñ'.

\begin{table}[h!]
\centering
\begin{tabular}{ll}
\toprule
Low tone           & e      \\
High tone          & \`{e}      \\
Falling tone       & \'{e}      \\
Low rising tone    & \^{e}      \\
Glottal (low rising tone) & e'      \\
Nasal vowel        & \underline{e} \\
Umlaut (lax) vowel & ë       \\ 
\bottomrule
\end{tabular}
\caption{Diacritics and special characters in Bribri}
\label{tab:diacritics-bribri}
\end{table}

In addition to appearing individually, the diacritics can be combined. For example, the word \textit{\`{\"{e}}} `only' is /\textipa{I\tone{55}}/, and it has a lax vowel with a falling tone. The word \textit{\underline{\'{e}}n} `liver' is /\~{e}\textipa{\tone{53}}/, and it contains a tense, nasal `e' with a falling tone.

Here we will use the \citet{cursobribri} orthography for the experiment, but other orthographies use other diacritics. For example \citet{gramaticajara} uses a tilde for nasals, and \citet{margery} uses ogoneks. More generally, there is wide variation in the orthographic representations of Bribri. For example, the word \textit{t\underline{aî}} `much' has been found in writing in 14 different ways.

\section{Evaluation of Different Approaches for Diacritic Restoration}

\subsection{Data Inputs}

In the diacritic restoration task, the input is a string without diacritics or non-alphabetical characters (e.g. Bribri \texttt{Is be shkena} `How are you?'), and the output is the string with its diacritics according to the community's orthographic norms (e.g. Bribri \texttt{\`{I}s be' shk\`{e}n\underline{a}}). Examples for both languages are shown in Table \ref{tab:example-inputs}. For the Bribri input we eliminated all diacritics and special characters from the vowels (tones, nasals, umlauts for lax vowels, apostrophe for glottal stops). For the CIM input we eliminated the diacritic macron from the vowels and the saltillo glottal stop consonant.

\begin{table}[h!]
\centering
\begin{tabular}{ll}
\toprule
Bribri Input & \texttt{Is be shkena}     \\
Output       & \texttt{\`{I}s be' shk\`{e}n\underline{a}}    \\
Meaning      & `How are you?'   \\ 
\midrule
CIM Input    & \texttt{i toku apii}      \\
Output       & \texttt{i t\={o}ku  {\textquotesingle}\={a}pi{\textquotesingle}i}     \\
Meaning      & `From my school' \\ 
\bottomrule
\end{tabular}
\caption{Examples of inputs}
\label{tab:example-inputs}
\end{table}

The Bribri dataset is made up of 10,962 sentences (78,784 words), which are part of the AmericasNLP dataset \citep{americsNLPGithub2021}. The CIM dataset contains 5,439 sentences (226,933 words) from the Te Vairanga Tuatua collection \citep{nicholas2012_tevairanga} in the Paradisec corpus \citep{thieberger2014paradisec}. We made 5 random train/valid/test partitions with 80\%, 10\% and 10\% of the words respectively. 

\subsection{Models}
We used three families of algorithms for our tests: character-based statistical machine translation (SMT) models, fine-tuned LLMs, and zero-shot LLMs. The SMT system Moses \citep{koehn-etal-2007-moses} was used as a baseline for all our experiments. SMT has been shown to be a strong baseline for diacritic restoration, and statistical methods remain useful in under-resourced languages settings \citep{kuparinen-etal-2023-dialect}. In addition, it is the closest approach to a rule-based system, which would require detailed knowledge of the languages in question, something not viable in all scenarios. We used the distribution by \citet{ljubesic-etal-2016-corpus,scherrer16-automatic} which runs Moses character-based models.\footnote{https://github.com/clarinsi/csmtiser} All Moses systems use a n-gram language model of size six.

The main LLM selected for fine-tuning was ByT5 \citep{xue-etal-2022-byt5}, chosen for two key reasons. First, it is a character-based model that operates directly on UTF-8 bytes, a feature that has proven effective for processing low-resource datasets. Second, it is multilingual, pretrained on the mC4 corpus \citep{colin-etal-2020-mC4}, which covers a wide range of languages. To enable a comparison with more conventional, token-based architectures, we also included the monolingual T5 \citep{JMLR:v21:20-074} and the multilingual mT5 \citep{xue-etal-2021-mt5}. We had to extend T5 and mT5's tokenizers to recognize Bribri diacritics by retraining SentencePiece \citep{kudo-richardson-2018-sentencepiece}. All three models were fine-tuned on our Bribri and CIM datasets and evaluated on their ability to restore diacritics.

For zero-shot LLM testing, we chose Claude Sonnet 4.5 2025-09-29 \citep{anthropic2025claudesonnet45}. We performed two types of experiments: First, we attempted a zero-shot diacritic restoration, where we gave the model the test sentences and asked it to add the diacritics. In the second experiment, we uploaded a portion of the training set (the equivalent of 150,000 tokens) and asked it to add the diacritics to the test sentences.



\subsection{Benchmarking Model Effectiveness}\label{section:results}

In order to evaluate performance, we trained, fine-tuned or prompted the algorithms using the training and validation sets, and evaluated using the word error rate (WER) by predicting the output from the test set inputs. We did this five times (for each of the randomly generated sets based on our total data) and calculated the average WER.

In addition to comparing the expected output with the predictions of each algorithm, we need to calculate the accuracy when trying to predict specific diacritics. In order to do this, we transform the output strings into strings that focus on a single type of diacritic. Table \ref{tab:error-transcriptions-bribri} shows an example for Bribri. In the \textit{Tones} transcription, each syllable is transcribed solely as its tone. Here, the first word \textit{\`{i}s} 'how' has a high tone, and is therefore represented as \texttt{H}. The second word has a glottal low rising tone, and is represented as \texttt{G}. In the \textit{Nasal} condition, the syllables are either oral (\texttt{O}) or nasal (\texttt{N}), and in the umlaut lax vowel condition, the syllables either have a tense (\texttt{T}) or a lax (\texttt{L}) vowel. These transformed strings allow for two calculations: (i) a measure of how many tonal, nasal or umlaut diacritics are wrong overall, and (ii) a measurement of how accurate is the marking for each algorithm. The first is reported using WER, and the second is reported using the F1 for each tone and diacritic. 

\begin{table}[h!]
\centering
\begin{tabular}{ll}
\toprule
            & Transcription \\ 
\cmidrule{2-2}
Translation & How are you doing, sir/ma'am?  \\ 
\midrule
All         & \texttt{\`{i}s be' shk\`{e}n\underline{a}, ak\'{\"{e}}k\"{e}pa} \\
Tones        & \texttt{H G HL LFLL}        \\
Nasal vowels & \texttt{O O ON OOOO}      \\
Umlaut (lax) & \texttt{T T TT TLLT}        \\ 
\bottomrule
\end{tabular}
\caption{Different transcriptions of the Bribri phrase \textit{How are you?} asked to an elder. These are used to analyze the different types of errors in the model. Tones distinguishes \{L:low, H:high, F:falling, R:rising, G:glottal\} tones. Nasal distinguishes between \{O:oral, N:nasal\} vowels. Umlaut distinguishes \{T:tense, L:lax\} vowels.}
\label{tab:error-transcriptions-bribri}
\end{table}

These transformations were also performed on the CIM text. In the \textit{Long/short vowel} condition, each syllable is represented by whether it has a long (\texttt{L}) or short (\texttt{S}) syllable. The \textit{Consonants and glottals} condition has three possibilities for each syllable: it has a glottal consonant (\texttt{G}), a non-glottal consonant (\texttt{C}), or it is only a vowel and has no consonant (\texttt{N}). A more simple version of this is the \textit{Presence of glottals} condition, which describes an entire word by the number of glottals it has (one or more \texttt{G}), or as having no glottals (\texttt{N}).

\begin{table}[h!]
\centering
\begin{tabular}{ll}
\toprule
            & Transcription \\ 
\cmidrule{2-2}
Translation & From my school  \\ \hline
All         & \texttt{i t\={o}ku  {\textquotesingle}\={a}pi{\textquotesingle}i} \\
Long and short vowels  & \texttt{S LS LSS}        \\
Consonants and glottals & \texttt{N CC GCG}      \\
Presence of glottals & \texttt{N N GG}        \\ 
\bottomrule
\end{tabular}
\caption{Different transcriptions of the CIM phrase \textit{the school} to analyze the different types of errors. Long/short distinguishes \{L: long, S: short\} vowels. Consonants has \{G: glottals, C: other consonants, N: words without glottals or consonants\}. Presense of glottals has \{G: glottal, N: word without glottals\}.}. 
\label{tab:error-transcriptions-cim}
\end{table}

\paragraph{Results}
The main baseline for our calculations is the ``no restoration" condition. These are the results if we compare the original text to a text without any diacritics at all, and it gives us an idea of the actual improvement for each algorithm. Table \ref{table:general-results} contains this and the results of the diacritic restoration for each algorithm. The baseline error rate is 85 for Bribri and 33 for CIM.

\begin{table}[h!]
\centering
\begin{tabular}{lcc}
\toprule
                 & Bribri & CIM \\ 
\midrule
\rowcolor{gray!40} No restoration   & 85     & 33  \\
Moses            & 88     & 38   \\
Claude Zero-Shot & 85     & 30  \\
Claude Few-Shot  & 40     & 11  \\
T5 Fine-tuning   & 32     & 9   \\
mT5 Fine-tuning  & 43     & 24   \\
ByT5 Fine-tuning & \textbf{16}     & \textbf{7}   \\ \bottomrule
\end{tabular}
\caption{Word Error Rate (WER) for statistical, fine-tuned LLM and zero-shot LLM approaches.}
\label{table:general-results}
\end{table}

The statistical and zero-shot LLM methods have generally worse results, and fine-tuning the character-based model have the best results. The statistical-based Moses has the largest error (Bribri: 88; CIM: 38). The LLM-based zero-shot using Claude also had very high error rates (Bribri: 85; CIM: 30), but showing examples to the LLM cuts the WER by half (Bribri: 40; CIM: 11). As for the T5 fine-tuning, the mT5 does not offer improvements in performance (Bribri: 43; CIM: 24), but the T5 does decrease the error (Bribri: 32; CIM: 9). The best result comes from fine-tuning ByT5, with WER of 16 for Bribri and 7 for CIM. These results are comparable to the ones for te reo M\={a}ori, while using a much smaller mass of data.

Table \ref{table:bribri-diacritics} shows the overall error for tonal, nasal and lax vowel diacritics in Bribri. Moses and the zero-shot Claude produced the results with the highest error. The few-shot Claude cut the error in half compared to the Zero-shot (72 versus 32). The mT5 and T5 fine-tuning had similar error rates; and again the ByT5 had the lowest error, with tones having only WER=13. Overall, tones are the most difficult diacritics to mark; the error rates for nasal and umlauts are 5 and 3 respectively.

\begin{table}[h!]
\centering
\begin{tabular}{lccc}
\toprule
                 & Tone & Nasal & Umlaut \\ 
\midrule
\rowcolor{gray!40} No restoration   & 69   & 34    & 24     \\
Moses            & 52   & 35    & 10     \\
Claude Zero-Shot & 72   & 40    & 30     \\
Claude Few-Shot  & 32   & 17    & 15     \\
T5 Fine-tuning   & 29    & 20     & 17      \\
mT5 Fine-tuning   & 38    & 28     & 24      \\
ByT5 Fine-tuning & \textbf{13}   & \textbf{5}     & \textbf{3}      \\ 
\bottomrule
\end{tabular}
\caption{WER for tonal, nasal and umlaut transcriptions in Bribri, as described in Table \ref{tab:error-transcriptions-bribri}.}
\label{table:bribri-diacritics}
\end{table}

The patterns observed in Bribri also held true for CIM. As can be seen in Table \ref{table:cim-specific-wer}, ByT5 has the lowest error rate. Interestingly, the few-shot Claude was closer in error reduction compared to the ByT5.

\begin{table}[h!]
\centering
\begin{tabular}{lcc}
\toprule
                 & \begin{tabular}[c]{@{}c@{}}Macron\\ (long vowel)\end{tabular} & \begin{tabular}[c]{@{}c@{}}Presence\\ of glottals\end{tabular} \\ 
\midrule
\rowcolor{gray!40} No restoration   & 21                                                            & 15                                                             \\
Moses            & 8                                                             & 15                                                              \\
Claude Zero-Shot & 19                                                            & 15                                                             \\
Claude Few-Shot  & 8                                                             & 2.8                                                            \\
T5 Fine-tuning   & 7                                                             & 4                                                              \\

mT5 Fine-tuning   & 19                                                             & 9                                                              \\

ByT5 Fine-tuning & \textbf{6}                                                             & \textbf{2.5}                                                            \\ 
\bottomrule
\end{tabular}
\caption{WER for macron and glottal transcriptions in CIM, as described in Table \ref{tab:error-transcriptions-cim}.}
\label{table:cim-specific-wer}
\end{table}

\paragraph{Confusion patterns in complete restoration}

We used the strings described in tables \ref{tab:error-transcriptions-bribri} and \ref{tab:error-transcriptions-cim} to determine how often there are tonal replacements (e.g. how often a correct high tone is replaced by an incorrect falling tone). Using these strings allows us to treat this problem as a classification problem and to calculate precision, recall, and F1. Table \ref{table:f1-bribri-diacritics} shows the F1 for specific diacritics in Bribri. Tonal diacritics that go directly on the vowels (i.e. high, falling and rising) have the lowest F1, and are therefore the most difficult for the computer to learn. Non-tonal diacritics, such as those that indicate nasal and lax vowels have relatively high performance, and vowels with no explicit tonal diacritics (low tone) have the best performance for all of the algorithms. The low performance of the zero-shot solution is noteworthy: all tonal and nasal diacritics have values below F1:23, which indicates that this method might not be suitable for this task.

\begin{table*}[h!]
\centering
\begin{tabular}{lccccccc}
\toprule
                 & \multicolumn{1}{l}{Tone} & \multicolumn{1}{l}{} & \multicolumn{1}{l}{} & \multicolumn{1}{l}{} & \multicolumn{1}{l}{} & \multicolumn{1}{l}{} & \multicolumn{1}{l}{} \\ \cmidrule{2-6}
                 & Low                      & High                 & Falling              & Rising               & Glottal              & Nasal                & Umlaut               \\\midrule
Moses            & 80                       & 55                   & 52                   & 54                   & 0                    & 0                    & 80                   \\
Claude Zero-Shot & 66                       & 20                   & 18                   & 2                    & 22                   & 23                   & 53                   \\
Claude Few-Shot  & 86                       & 59                   & 53                   & 42                   & 83                   & 76                   & 75                   \\

T5 Fine-tuning   & 81                        & 56                    & 50                    & 48                    & 74                    & 66                    & 64                    \\

mT5 Fine-tuning   & 79                        & 53                    & 44                    & 43                    & 64                    & 64                    & 62                    \\

ByT5 Fine-tuning & \textbf{96}              & \textbf{85}          & \textbf{80}          & \textbf{78}          & \textbf{93}          & \textbf{94}          & \textbf{95}          \\
\bottomrule
\end{tabular}
\caption{Average F1 for individual diacritics for Bribri.}
\label{table:f1-bribri-diacritics}
\end{table*}

Table \ref{table:f1-cim-diacritics} shows the F1 for diacritics in CIM. The ByT5 has F1 results of 89 or above. The few-shot Claude is the runner-up, but its performance is 23 points lower for glottals. In general, glottals appear to be difficult for all algorithms except for the ByT5.

\begin{table}[h!]
\centering
\begin{tabular}{lcc}
\toprule
                 & \multicolumn{1}{l}{Macron} & \multicolumn{1}{l}{Glottal} \\ 
\midrule
Moses            & 77                          & 26                           \\
Claude Zero-Shot & 40                         & 12                          \\
Claude Few-Shot  & 78                         & 68                          \\
T5 Fine-tuning   & 71                          & 58                           \\
mT5 Fine-tuning   & 39                          & 37                           \\
ByT5 Fine-tuning & \textbf{89}                & \textbf{91}                 \\ 
\bottomrule
\end{tabular}
\caption{Average F1 for CIM diacritics}
\label{table:f1-cim-diacritics}
\end{table}

\subsection{Amounts of Data and Performance}\label{sec:datamass}

Once we determined the algorithm with the best performance, we created a series of smaller subsets of the training and validation data, using randomly selected sentences from the main train/validation sets. We then calculated the WER when the algorithm was trained on these lower masses of data. The objective of this experiment is to determine how much data is required to reach certain levels of error rates. This can be used by researchers and community members to determine whether training these diacritic restoration algorithms is possible with the available data for their languages.

For Bribri, we created five subsets with \{1000, 5000, 10000, 25000, 50000\} words each. For CIM, we created five subsets with \{1000, 5000, 10000, 25000, 50000, 100000\} words each. We used these to train and evaluate ByT5 models.

\paragraph{Results}

\begin{figure}[!ht]
\begin{center}
\includegraphics[width=\columnwidth]{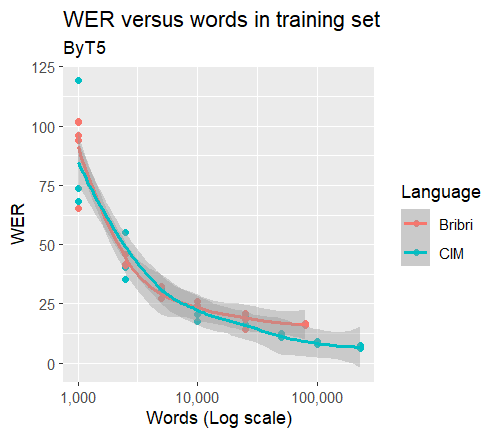}
\caption{WER of ByT5 outputs as a function of word presence in the training set.}
\label{fig-data-mass}
\end{center}
\end{figure}

Figure \ref{fig-data-mass} shows the relationship between number of words in the dataset and word error rate. There is significant variation with low data masses, e.g., 1000 words; approximately 2 pages of single-spaced text in English. At this point, the expected performance is about WER=80-90 for both languages. A performance of about WER=25 is reached with a mass of approximately ten thousand words (about 20 pages of single-space text). After ten thousand words, the decrease in error rate becomes slower and more linear.

\subsection{Comparison with High-Resource Languages}

In order to better understand the performance of the best algorithm on low-resource languages, we also trained models using datasets from three higher-resourced languages: Spanish, Vietnamese and Pinyin-transcribed Mandarin Chinese. 

We selected Spanish as a comparison with CIM because Spanish vowels also have two diacritics (the acute accent \texttt{é} and the umlaut \texttt{\"{u}}). We used the MLS dataset \citep{pratap2020mls} and extracted a dataset of 6325 sentences (226929 words), similar in size to the CIM sample.

In order to create a comparison to Bribri, we selected Vietnamese and Mandarin transliterated into Pinyin. Vietnamese was used because it is the most directly comparable high-resource language to Bribri: It is the only language that is higher-resourced and has tonal diacritics in the Roman alphabet. In addition to this, Vietnamese is a language that is included in the mC4 dataset, and Vietnamese and Bribri have similar diacritic densities: In the Vietnamese corpus, 27\% of the characters are diacritics, while in the Bribri corpus, 31\% of the characters are diacritics. There are some differences: Vietnamese has nine diacritics for different tones and vowel qualities, higher than the number of diacritics in Bribri (6). We selected a set of 7716 sentences (78793 words) from the VIVOS corpus \citep{luong2016non} and split it into five random train/valid/test sets.

As for Mandarin, its Pinyin representation is also comparable to Bribri in several ways. It has four tonal diacritics and one umlaut to indicate a rounded vowel. Like Bribri, it is not included in pretraining sets, as Mandarin is usually represented using Chinese characters. Also, Pinyin has a lower but similar diacritic density (22\% of the characters are diacritics, compared to Bribri's 30\%). We selected a set of 7244 sentences (78779 words) from the Aishell-3 corpus \citep{aishellshi2020} and again split it into five random train/valid/test sets.

\paragraph{Results}

Table \ref{table:big-langs-byt5} shows the WER for the low-resource languages (i.e. Bribri, CIM) and the high-resource languages chosen as comparison points (i.e. Spanish, Vietnamese and Pinyin-transliterated Mandarin Chinese).

\begin{table}[h!]
\centering
\begin{tabular}{lcc}
\toprule
                 & \begin{tabular}[c]{@{}c@{}}No\\ restoration\end{tabular} & \begin{tabular}[c]{@{}c@{}}ByT5\\ Fine-tuning\end{tabular} \\ \midrule
Bribri           & 85                                                       & 16                                                         \\
Mandarin (Pinyin) & 94                                                       & 22                                                         \\
Vietnamese       & 87                                                       & 16                                                         \\ \midrule
CIM              & 33                                                       & 7                                                          \\
Spanish          & 12                                                       & 3                                                          \\ 
\bottomrule
\end{tabular}
\caption{Average WER diacritic restoration in high and low-resource languages}
\label{table:big-langs-byt5}
\end{table}

ByT5 reached a similar WER values in Vietnamese as it did in Bribri, but this might not mean that the learning process was the same. On one hand, Vietnamese does appear in the ByT5 pretraining set, which might give it a leg up. On the other hand, the Vietnamese diacritic system is more complex. Not only does Vietnamese have more diacritics (9 versus Bribri's 6), it also has a higher tonal density and the tones themselves have a higher functional load \citep{hockett1967quantification}. This means that they distinguish more minimal pairs, making them more difficult to tag correctly. In contrast, Bribri has fewer diacritics per word, only one per morpheme, and its root morphemes tend to be polysyllabic. Therefore, even if both Vietnamese and Bribri have a WER=16, the results for Vietnamese are better because it had to solve a more difficult problem. Notice also that, due to the low mass of Vietnamese data it is seeing, these results are naturally lower than those for previous work \citep{le2021diacritics,stankevivcius2022correcting}.

As for Pinyin, its performance was lower than the one for Bribri and Vietnamese. This is to be expected, as (i) it has numerous diacritics, (ii) it has a high density of text with diacritics, and the diacritics carry a high functional load, and (iii) the Mandarin text that ByT5 comes pretrained with possibly includes very little Pinyin, and even less text with Pinyin tones, as this text rarely occurs in public spaces \citep{mathias1980computers}.


In the case of Spanish, the restoration of diacritics is aided by the knowledge in the pretraining. Both CIM and Spanish see their error reduced to a quarter by the fine-tuning, but the error rate for Spanish is very low, WER=3.

\subsection{Diacritic Correction}

In our final experiment, we replicate spell checking, where some diacritics might be correct but some might not be. We create synthetic ``wrong" inputs by introducing probabilistic constraints on the appearance of diacritics, based on common replacements observed in naturalistic text. For example, in Bribri, if the original string has a vowel with a high tone, we set a 15\% probability that the tone will be represented as a falling tone, 15\% probability that it will stay as a high tone, and a 70\% probability that the tone diacritic will be absent.

This procedure was also used to created CIM strings with synthetic spelling mistakes. For example, if a word starts with the causative prefix \{{\textquotesingle}aka-\}, we set a 20\% probability that the glottal stop will appear. If the word ends with the nominalizer suffix \{-{\textquotesingle}anga\}, there's a 50\% probability that the glottal stop will appear. In other cases, if the string has a glottal stop, there is a 20\% chance that this will appear in the synthetic text.

We used this transformation to modify the train/valid/test sets for each language. These were used to train models using the best-performing algorithm. The ultimate objective of this experimental condition is to replicate spell checking, and study how these systems might behave when correcting real texts with spelling variation.

\paragraph{Results}

The experiment for the simulated spelling correction had results of WER=7 for CIM and WER=13 for Bribri. The results for Bribri are slightly better than those for the complete restoration condition (WER=16), but they were the same for CIM. More research is needed, in particular, the building of a corpus of natural Bribri and its corrected version in order to further study this problem.

\section{Discussion}


\subsection{Classical versus LLM-based work on Under-Resourced Languages}
Whether classical machine learning or LLM-based approaches are "better" for under-resourced languages depends greatly on the definition of "under-resourced" and the goals of the project. For extremely low-resource languages that need immediate work from language documenters, fine-tuning an LLM or using an adequately controlled commercial LLM might make sense for some tasks. When there is a similar or related language in the pretraining data, for example, this might help. This was observed for the zero and few shot conditions: CIM had much lower error rates than Bribri, possibly because the models include data for other Polynesian languages. In addition to this, if the task can be accommodated to the point where specialized knowledge of the language is not required (e.g., diacritic restoration), these solutions will be more efficient than creating a rule-based system, once a certain mass of data is reached. On the other hand, maintaining control of hallucinations made the task complicated. For example, only small amounts of data could be processed at a time; giving the system large lists would create confusion in its outputs.

Classical approaches, however, remain more accessible from a computational perspective. For instance, our Moses models ran entirely on CPUs. Nevertheless, we observed substantially lower WER and higher F1 performance with byT5, which is also a character-level model. This indicates that knowledge encoded during pre-training can be leveraged for under-resourced languages. While Moses did not achieve high overall performance, it did particularly well with umlaut and macron transcription. This suggests a strong sensitivity to context-dependent phenomena, as expected from an n-gram language model, and indicates that larger data amounts may be required for the SMT model to reach its full potential. 

ByT5 clearly outperformed all other models, reinforcing the understanding that tokenization continues to pose challenges for NLP. Its description paper actually positions it as a step towards tokenizer-free approaches.  Because ByT5 decomposes each character into separate utf-8 bytes, it can learn fine-grained character-level patterns such as those found in Bribri. Off-the-shelf tokenizers, in contrast, are trained mainly on high-resource languages, needing re-training in the context of under-resourced languages like those presented here. This process can introduce biases from vocabulary size or BPE merges when working with a new language. The trade-off is that ByT5 has higher memory and compute requirements, and it is an open question how well this approach scales up to typical NLP tasks beyond the character-level transduction experiments reported here.



\subsection{Multilingual Models}

Multilingual LLMs are trained under the assumption that exposure to multiple languages help generalization, an approach that has shown success in other tasks such as machine translation \citep{nllbteam2022languageleftbehindscaling}. However, under-resourced languages vary widely in their amount of available digital data \citep{joshi2020state}. As a result, although under-resourced languages may benefit from the transfer of knowledge from the large-resourced ones, this is not guaranteed. A typologically related language may not be present in the pretraining data, and the quality of Indigenous language data might be low because it has been gathered without the collaboration or consent of the community. 


In our experiments, we observe that multilingual word-level models did not outperform their monolingual counterparts. When benefits from multilingual pretraining emerged, they were limited to the high-resource languages in its collection (e.g., Spanish). The multilingual pretraining itself did not seem to be aiding Bribri or CIM, given the low performance of the mT5 models and the Claude experiments. These findings suggest that, while LLMs can be useful for fine-tuning, the most effective base models may not be massively multilingual ones. Instead, models specialized or adapted for a single under-resourced language may yield better results.

\subsection{Tones}

Tones are very difficult to transcribe reliably, even for highly qualified and experienced linguists and community members. This also appears to be true of computer systems. The tonal languages had the lowest overall performance, and the tonal diacritics for Bribri were its lowest performing ones. A tool that can aid with tonal transcription is highly necessary for documentary linguists, and this is an underexplored area in the literature. We encourage more researchers to explore this problem.


\subsection{Data Sovereignty Concerns}

Not only did we have issues with the computing power used for the fine-tuning of LLMs, we also had to be careful with data sovereignty issues while using our data. While all of the raw data we used is available through public sources, its labeled form is not. We used Claude because it allowed privacy options so that the datasets would not be incorporated in future models, an option that wasn't available with models from OpenAI \citeyearpar{openai_chatgpt_2025}, for example.

While the use of this tool might be desirable to specialists and people working on language revitalization, there are potential risks in releasing them to the public. Releasing it to the public might increase interest in learning the language amongst community members, but it could also result in an unintentional endorsement of a certain way of writing these languages. Not only might this be biased towards one language variety or another, but it might be biased towards LLM hallucinations. These might be construed to be more authoritative given the elevated status of computer systems in 21st century life \citep{eubanks2025automating}, and might be considered the ``correct" form of these languages, intimidating actual speakers who speak correctly but who feel that their correct speech and writing diverges from the computer's recommendations.



\section{Conclusions}

In this paper we have explored the task of diacritic restoration on two extremely under-resourced languages. We found that some Deep Learning methods, i.e., character-based LLMs, are beneficial in a variety of sub-tasks in this setting, even with very small amounts of data. We also explored how these algorithms work with tonal languages, whose transcription is vexing and time-consuming for linguists and community members who work on language documentation.




\section{Bibliographical References}\label{sec:reference}

\bibliographystyle{lrec2026-natbib}
\bibliography{lrec2026-example}


















\end{document}